\RequirePackage{fix-cm}
\documentclass[smallextended]{svjour3}       
\smartqed  
\usepackage{graphicx}%
\usepackage{multirow}%
\usepackage{amsmath,amssymb,amsfonts}%
\usepackage{mathrsfs}%
\usepackage[title]{appendix}%
\usepackage{xcolor}%
\usepackage{textcomp}%
\usepackage{manyfoot}%
\usepackage{booktabs}%
\usepackage{algorithm}
\usepackage{algorithmic}
\usepackage{caption}
\usepackage{subcaption}
\usepackage{listings}%
\usepackage{float}

\graphicspath{{./images/}}
\begin{document}

\title{Dynamic Meta-Layer Aggregation for Byzantine-Robust Federated Learning}


\author{Reek Das \and Biplab Kanti Sen 
}


\institute{Reek Das \at
              APC Roy Government College \\
              \email{reekdas34@gmail.com}           
           \and
         Biplab Kanti Sen \at
         Assistant Professor,
         Department of Computer Science\\
              P.R. Thakur Government College\\
              \email{bksen.cu@gmail.com}
}

\date{Received: date / Accepted: date}

\maketitle

\begin{abstract}
Federated Learning (FL) is increasingly applied in sectors like healthcare, finance, and IoT, enabling collaborative model training while safeguarding user privacy. However, FL systems are susceptible to Byzantine adversaries that inject malicious updates, which can severely compromise global model performance. Existing defenses tend to focus on specific attack types and fail against untargeted strategies, such as multi-label flipping or combinations of noise and backdoor patterns. To overcome these limitations, we propose FedAOT—a novel defense mechanism that counters multi-label flipping and untargeted poisoning attacks using a metalearning-inspired adaptive aggregation framework. FedAOT dynamically weights client updates based on their reliability, suppressing adversarial influence without relying on predefined thresholds or restrictive attack assumptions. Notably, FedAOT generalizes effectively across diverse datasets and a wide range of attack types, maintaining robust performance even in previously unseen scenarios. Experimental results demonstrate that FedAOT substantially improves model accuracy and resilience while maintaining computational efficiency, offering a scalable and practical solution for secure federated learning.
\end{abstract}

\section{Introduction}\label{sec1}
Federated Learning (FL) has emerged as a transformative paradigm for distributed model training, enabling multiple clients to collaboratively learn a global model without sharing their raw data \cite{mcmahan2017,konevcny2016federated}. This decentralized framework is particularly crucial for privacy-sensitive domains such as healthcare, finance, and mobile applications, where data confidentiality and regulatory compliance are essential \cite{yang2019federated,li2020federated}. By transmitting only model updates to a central server, FL preserves data locality while significantly mitigating privacy risks and ensuring compliance with key data protection regulations such as the General Data Protection Regulation (GDPR) and the Health Insurance Portability and Accountability Act (HIPAA) \cite{truong2021privacy}.

However, the distributed and client-driven architecture of FL introduces new vulnerabilities that do not appear in centralized learning settings. Among various security threats in federated learning, Byzantine attacks, in which certain clients act maliciously or become compromised, represent one of the most severe challenges \cite{blanchard2017,shi2025byzantine}. Byzantine clients can inject poisoned gradients, flip labels, or manipulate updates to mislead the global aggregation process and degrade model convergence. These threats become more damaging when the data across clients are non-IID (non-independent and identically distributed), meaning that each client holds data with differing statistical properties or class imbalances. Under such heterogeneous conditions and variable client participation, where different subsets of clients take part in each training round due to network or resource constraints, the global model already struggles to learn consistent patterns, and malicious updates further compromise the stability and reliability of the learning process \cite{yin2018,guerraoui2018}.

Several aggregation strategies have been developed to enhance the robustness of federated learning against Byzantine threats. Classical approaches such as Krum \cite{blanchard2017}, Trimmed Mean \cite{yin2018}, and Bulyan \cite{guerraoui2018} attempt to statistically filter or weight client updates based on assumptions about the proportion and behavior of adversaries. More recent methods, including FoolsGold \cite{fung2020limitations}, RobustFedAvg \cite{pillutla2022robust}, and Byzantine-Robust Decentralized Federated Learning \cite{he2021byzantine}, have explored geometric median aggregation, reputation-based adjustment, or peer-consensus mechanisms to resist poisoned updates. Although these approaches have advanced the field, their robustness often depends on static thresholds, pre-defined statistical bounds, or specific assumptions about the attack model. As a result, their performance degrades in adaptive or mixed attack scenarios, where adversaries vary their strategies across training rounds to evade detection \cite{baruch2019,xie2020}.

Recent research, such as Shi et al. (2025) \cite{shi2025byzantine}, has further investigated gradient similarity–based methods to identify and isolate malicious clients. Building on this growing body of work, our study introduces a new perspective through \textbf{FedAOT (Federated Adaptive Optimal Tuning)}, a meta-learning–driven defense framework designed to enhance resilience against untargeted and multi-label poisoning attacks.

The main contributions of this work are summarized as follows:

\begin{itemize}
    \item A meta-learning–based aggregation strategy that adaptively assigns client weights using validation feedback to enhance robustness.  
    \item An adaptive optimization mechanism that reduces the impact of unreliable or adversarial updates without prior attack assumptions.  
    \item A unified defense against untargeted poisoning and label-flipping attacks under non-IID and heterogeneous data distributions. 
    \item Empirical evaluation demonstrating superior resilience and convergence over existing Byzantine-robust approaches.  
\end{itemize}

Overall, FedAOT underscores the potential of meta-learning as a practical and scalable approach to strengthening the robustness of federated learning systems. By introducing adaptivity into the aggregation mechanism, it moves beyond static, rule-based defenses toward more intelligent and self-adjusting learning paradigms. 
The remainder of this paper is organized as follows: Section~\ref{relatedwork} reviews related Byzantine-resilient FL methods; Section~\ref{methodology} details the proposed FedAOT framework; Section~\ref{experiments} presents experimental evaluations and results; and Section~\ref{conclusion} concludes the paper with discussions on implications and future research directions.
(Note: Since multi-label flipping and untargeted poisoning attacks operate similarly in the context of label flipping, we will refer to both as 'untargeted poisoning attacks' throughout this paper for clarity.)

\section{Related Work}\label{relatedwork}
Federated Learning (FL) distributes model training across multiple clients, enabling privacy-preserving computation but introducing vulnerabilities to Byzantine attacks, where malicious participants manipulate local updates to degrade the global model. This challenge has led to extensive research on robust aggregation and adaptive defense mechanisms.

Early aggregation strategies, such as \textit{Federated Averaging (FedAvg)}~\cite{mcmahan2017}, simply average local model updates. Although efficient in benign settings, FedAvg performs poorly when even a small fraction of clients behave maliciously, since poisoned updates are directly incorporated into the global model. To counter this, \textit{Krum}~\cite{blanchard2017machine} selects the update most consistent with others, assuming that the majority of clients are honest. While it provides basic robustness, Krum’s performance deteriorates when adversarial clients constitute a large fraction or when data heterogeneity causes honest updates to diverge significantly. Similarly, \textit{Geometric Median (GeoMed)}~\cite{yin2018byzantine} replaces the arithmetic mean with a median-based aggregation to suppress outliers, but it suffers when attackers generate carefully crafted updates that remain statistically indistinguishable from benign gradients.

Other methods, such as \textit{Foolsgold}~\cite{fung2020limitations}, analyze the similarity of client gradients to identify potential collusion among malicious agents. Although effective against targeted attacks, Foolsgold performs inconsistently under untargeted poisoning scenarios where attackers act independently. Later approaches like \textit{FABA (Filtering Adversaries By Agreement)}~\cite{li2019faba} and \textit{BRIEF (Byzantine-Robust Federated Learning via Optimal Voting)}~\cite{wang2022brief} refine this concept by adaptively filtering updates based on mutual agreement or voting mechanisms. However, these algorithms still depend on static heuristics or assumptions about attack types, which limits their adaptability to new or evolving threat patterns. Furthermore, most of these techniques rely on fixed similarity thresholds or require a clean reference model, conditions rarely available in real-world federated systems.

To address these limitations, researchers have proposed adaptive and learning-guided defenses. \textit{FLTrust}~\cite{cao2021fltrust} introduces a server-side trusted dataset to calibrate client updates, achieving good robustness under controlled environments but facing scalability issues in privacy-sensitive deployments. \textit{RobustFL}~\cite{zhang2022robustfl} assigns confidence-based weights to updates by modeling uncertainty, enhancing stability under mixed attack scenarios. Meanwhile, personalized FL frameworks~\cite{fallah2020personalized} explore fine-grained, feedback-based adaptation of model parameters. These works have inspired dynamic aggregation strategies that move beyond static heuristic rules.

Building on these foundations, \textit{Shi et al. (2025)}~\cite{shi2025byzantine} proposed the \textit{RSDFL} framework, which measures pairwise gradient similarity to identify and isolate Byzantine clients. RSDFL significantly improves resilience against direct poisoning but assumes consistent gradient behavior across clients and relies on static similarity thresholds. Consequently, its robustness declines under heterogeneous data distributions or when attack strategies evolve dynamically over training rounds.

Considering a related but distinct perspective, \textit{LoRA (Low-Rank Adaptation)}~\cite{hu2021lora} fine-tunes a limited set of low-rank parameters, typically within the final layer. Recent trends have focused on meta-learning and reinforcement learning-based aggregation, aiming to continuously adapt aggregation behavior based on client reliability feedback. \textit{FedRAD}~\cite{li2023fedrad} employs reinforcement learning to optimize aggregation policies dynamically, showing improved generalization under dynamic attack patterns. \textit{TrustFed}~\cite{wang2024trustfed} introduces a historical trust modeling approach, where each client’s contribution is weighted by a dynamically updated reliability score. These frameworks mark a transition toward self-optimizing FL systems that can learn to defend rather than rely on static defense parameters.

Despite this progress, existing Byzantine-resilient frameworks share key limitations: (1) their reliance on predefined heuristics or trust metrics restricts adaptability, (2) their defensive behavior is often reactive rather than predictive, and (3) their effectiveness declines when adversaries evolve or when honest client updates temporarily degrade due to data heterogeneity. These challenges motivate the need for a learning-driven, feedback-aware aggregation mechanism that can continuously refine its defense strategy based on observed model performance. This motivation forms the foundation of the proposed FedAOT framework.

\section{Problem Formulation and Motivation}
Federated Learning (FL) enables decentralized model training across a set of distributed clients 
$\{C_1, C_2, \dots, C_N\}$, where each client $C_i$ holds a private dataset $D_i$ and performs 
local optimization without sharing raw data. The global learning objective is formulated as:
\[
\min_{w} \; F(w) = \sum_{i=1}^{N} p_i F_i(w),
\]
where $F_i(w) = \mathbb{E}_{(x,y) \sim D_i} [\ell(w; x, y)]$ denotes the local empirical loss on client $i$, 
$p_i = \frac{|D_i|}{\sum_j |D_j|}$ is the data proportion weight, and $\ell(\cdot)$ is the task-specific loss function.
At each communication round $t$, clients compute local updates $\Delta w_i^{(t)}$ and send them to the server, 
which aggregates them to update the global model:
\[
w^{(t+1)} = w^{(t)} + \eta \cdot \text{Agg}\left( \{ \Delta w_i^{(t)} \}_{i=1}^N \right),
\]
where $\eta$ is the global learning rate and $\text{Agg}(\cdot)$ denotes the aggregation function.

\subsection*{Byzantine Challenge:} In realistic federated environments, a subset of clients may act maliciously by injecting poisoned or corrupted updates. Let $\mathcal{A} \subset \{1, \dots, N\}$ denote the set of Byzantine clients, each contributing adversarial updates $\Delta w_i^{\mathcal{A}} \sim \mathcal{D}_{\text{adv}}$. 
These updates are often statistically similar to benign ones:
\[
\mathbb{E}[\Delta w_i^{\mathcal{A}}] \approx \mathbb{E}[\Delta w_j], \quad \forall j \notin \mathcal{A},
\]
making them difficult to detect through conventional distance or similarity-based filtering. Most existing Byzantine-resilient methods focus on \emph{targeted poisoning attacks}, where adversaries manipulate 
updates to misclassify specific classes. However, \emph{untargeted model poisoning} poses an equally severe yet 
harder-to-detect threat. In such cases, malicious clients inject random noise or indiscriminate label flips, 
degrading global model performance without a specific target. These noisy updates often mimic natural client 
variations under non-IID conditions, allowing them to bypass standard defense mechanisms. Consequently, traditional aggregation schemes, whether update-based or model-based, often fail to converge:
\[
\begin{aligned}
w^{(t+1)} &= w^{(t)} + \eta \cdot \text{Agg}(\{\Delta w_i^{(t)}\}_{i=1}^N)
&&\text{fails to converge as } t \to \infty, \\[0.5em]
w^{(t+1)} &= \text{Agg}(\{\Delta w_i^{(t)}\}_{i=1}^N)
&&\text{\shortstack[l]{may oscillate or stagnate \\ under adversarial influence.}}
\end{aligned}
\]

This challenge is further amplified in heterogeneous and resource-constrained edge environments, 
where client data distributions and participation patterns vary significantly.
\vspace{-0.1em}
{Motivation}
To overcome these challenges, this work proposes a \textbf{robust and adaptive aggregation mechanism} 
that dynamically learns to assign client-specific weights based on their contribution reliability. 
The aggregation process is reformulated as:
\[
w^{(t+1)} = \sum_{i=1}^{N} k_i^{(t)} \Delta w_i^{(t)}, 
\quad \text{where } k_i^{(t)} \in [0, 1] \text{ and } \sum_{i=1}^{N} k_i^{(t)} = 1.
\]
Unlike heuristic or threshold-based defenses, the proposed approach adaptively infers $k_i^{(t)}$ 
using validation feedback, allowing it to suppress harmful updates and amplify trustworthy ones. 
This adaptive learning of aggregation weights aims to ensure robustness, convergence stability, 
and computational efficiency even under untargeted or evolving Byzantine threats in FL settings.
\\
\\

\begin{figure}[htbp]
\centering
\includegraphics[width=0.85\linewidth]{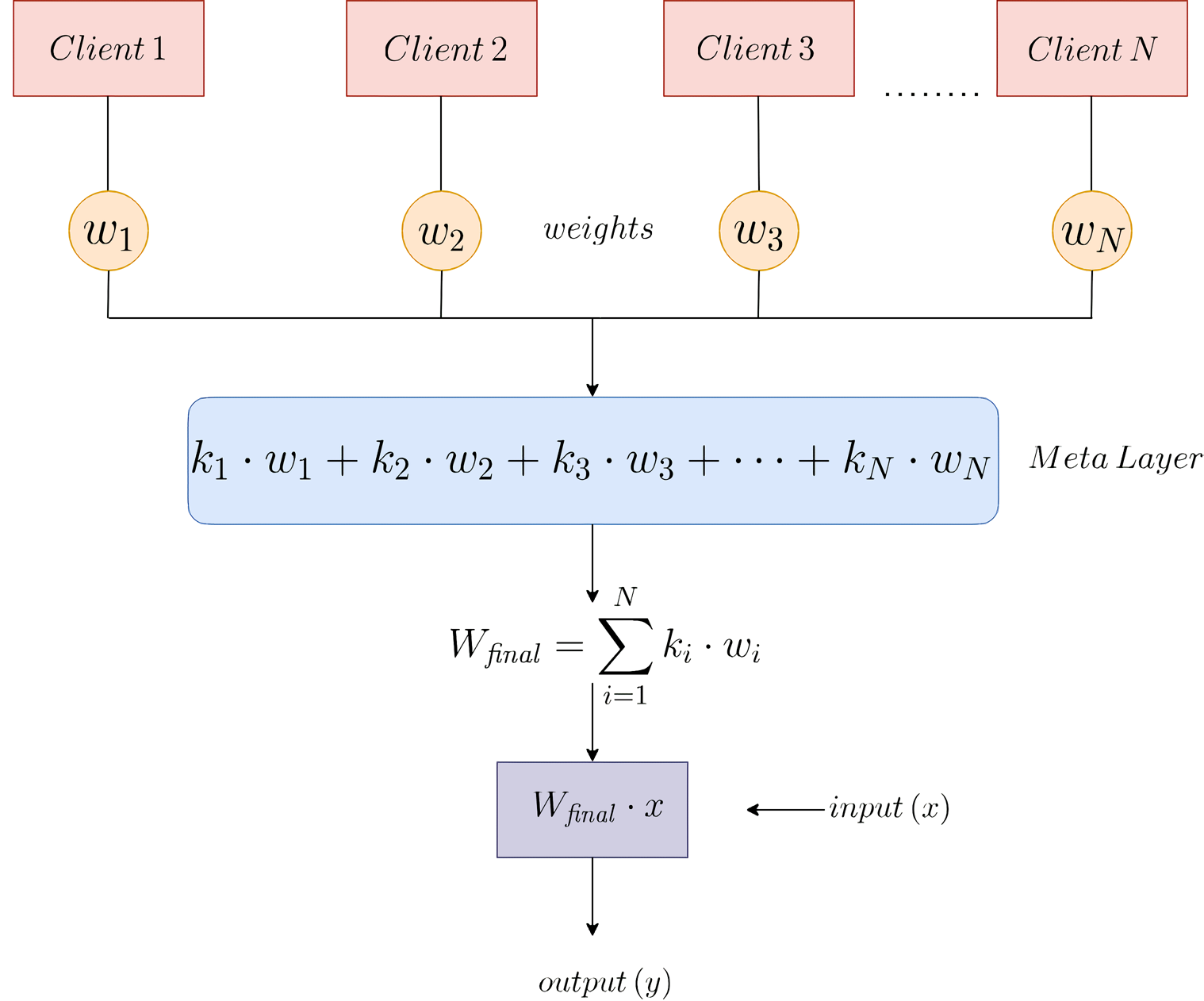}
\caption{Overview of the proposed adaptive meta-layer aggregation mechanism. Each client $C_i$ produces a local model $w_i$, and the meta-layer learns client-specific weights $k_i$ to compute a robust aggregated model $W_{\text{final}} = \sum_{i=1}^{N} k_i w_i$.}
\label{fig:meta_layer}
\end{figure}

\section{Proposed Methodology}\label{methodology}
Federated Adaptive Optimal Tuning (FedAOT) augments the standard server aggregation in federated learning with a lightweight, server-side meta-layer that learns client-specific importance weights. At each communication round, the server collects local updates from participating clients, composes a weighted aggregation using the current importance vector, evaluates the aggregated model on a small held-out meta-validation set, and applies a meta-gradient step to adjust the importance weights so as to reduce the validation loss. This adaptive re-weighting is intended to suppress the influence of adversarial or noisy updates while preserving contributions from helpful clients, and it incurs only a modest server-side overhead compared to local client computation.

\begin{algorithm}[t]
\caption{FedAOT: Federated Adaptive Optimal Tuning}
\label{alg:fedAOT}
\begin{algorithmic}[1]
\REQUIRE Clients $\{C_1,\ldots,C_N\}$; initial global model $W^{(0)}$; meta learning rate $\eta$; server meta-validation set $(x,y)$
\ENSURE Final aggregated global model $W_{\text{final}}$

\STATE \textbf{Initialize:} Set importance weights $k_i^{(0)} = \frac{1}{N}$ for all $i$

\FOR{communication round $t = 1$ to $T$}
\STATE Server broadcasts \(W^{(t)}\) to selected clients
    \STATE \textbf{Client-side Local Training}
    \FOR{each client $C_i$ in parallel}
        \STATE Client $C_i$ performs local training and returns update $\omega_i^{(t)}$
    \ENDFOR

    \STATE \textbf{Server Aggregation}
    \STATE \hspace{10 em} $W^{(t)} \gets \sum_{i=1}^{N} k_i^{(t)} \cdot \omega_i^{(t)}$
    \STATE \textbf{Meta-layer Update}
    \STATE Compute validation prediction: $\hat{y} \gets f(W^{(t)}, x)$
    \STATE Compute meta-loss: $L \gets \mathrm{Loss}(\hat{y}, y)$

    \FOR{each client $i$}
        \STATE Compute gradient: $g_i \gets \nabla_{k_i^{(t)}} L$
        \STATE Update importance weight: 
            $k_i^{(t+1)} \gets k_i^{(t)} - \eta \cdot g_i$
    \ENDFOR

    \STATE \textbf{Stabilization of Client Importance Weights}

    \IF{Normalization (used in this work)}
        \FOR{each client $i$}
            \STATE \hspace{7 em} $k_i^{(t+1)} \gets 
            k_i^{(t+1)} \,/\, 
            \sum_{j=1}^{N} k_j^{(t+1)}$
        \ENDFOR
    \ELSE
        \STATE \textbf{(better option) SoftMax-based stabilization}
        \FOR{each client $i$} \vspace{0.5 em}
            \STATE \hspace{4 em} $k_i^{(t+1)} \gets 
            \exp(k_i^{(t+1)}) \,/\, 
            \sum_{j=1}^{N} \exp(k_j^{(t+1)})$
        \ENDFOR
    \ENDIF
\ENDFOR
\STATE \textbf{Return} $W_{\text{final}} \gets W^{(T)}$
\end{algorithmic}
\end{algorithm}
 
\begin{table}[h]
\centering
\caption{Notation used in the FedAOT description.}
\label{tab:notation_fedaot}
\begin{tabular}{ll}
\toprule
Symbol & Description \\
\midrule
\(N\) & Total number of clients in the federation \\
\(C_i\) & Client indexed by \(i\) \\
\(W^{(t)}\) & Global model parameters at round \(t\) \\
\(\omega_i^{(t)}\) & Client \(i\)'s update at round \(t\) (model difference or gradient) \\
\(k_i^{(t)}\) & Importance weight for client \(i\) at round \(t\) (\(k_i^{(0)}=1/N,\; k_i^{(t)}\ge 0\)) \\
\(s_i^{(t)}\) & Optional logit for SoftMax parametrization of \(k_i\) \\
\(\eta\) & Meta learning rate for importance updates (server-side) \\
\((x,y)\) & Server-side meta-validation dataset (held-out) \\
\(L\) & Meta-validation loss evaluated on \((x,y)\) \\
\(\tau\) & SoftMax temperature (optional) \\
\(\alpha\) & Exponential smoothing coefficient (optional) \\
\(\varepsilon\) & Clipping lower bound for stabilized weights (optional) \\
\bottomrule
\end{tabular}
\end{table}

The central mechanism of FedAOT proceeds as follows. After each communication round the server forms the aggregated model
\[
W^{(t)} = \sum_{i=1}^N k_i^{(t)}\,\omega_i^{(t)},
\]
where \(\{\omega_i^{(t)}\}\) are the client updates collected that round and \(\{k_i^{(t)}\}\) are the current importance weights. The server evaluates the aggregated model on a small held-out meta-validation set \((x,y)\) and computes the meta-validation loss \(L=\mathrm{Loss}(f(W^{(t)},x),y)\). The sensitivity of this loss to each importance weight is given by the meta-gradient \(g_i^{(t)}=\nabla_{k_i}L\). Importance values are updated by a small meta-step
\[
k_i^{(t+1)} \gets k_i^{(t)} - \eta\, g_i^{(t)},
\]
and then stabilized to enforce a valid convex weighting for aggregation.

In our experiments, we used simple renormalization (division by the sum) as the default stabilization:
\[
k_i \leftarrow \frac{k_i}{\sum_j k_j},
\]
because it is computationally trivial and, together with conservative meta learning rates and Adam-based local training, produced stable behaviour across benchmarks. In practice we initialize \(k_i^{(0)}=1/N\) and enforce \(k_i^{(t)}\ge 0\).

For deployments facing extreme heterogeneity or noisy meta-gradients, we recommend the SoftMax parametrization on internal logits \(s_i\):
\[
k_i = \frac{\exp(s_i/\tau)}{\sum_{j=1}^N \exp(s_j/\tau)},\qquad \tau>0,
\]
and update \(s_i\) by meta-gradient descent. SoftMax guarantees \(0<k_i<1\) and avoids negative or extremely large intermediate values; use it when raw meta-gradients are noisy or when numerical stability is a concern.

Optionally apply exponential smoothing
\[
\tilde{k}_i^{(t+1)} = \alpha\, k_i^{(t)} + (1-\alpha)\, k_i^{(t+1)}
\]
and clip \(\tilde{k}_i^{(t+1)}\in[\varepsilon,1]\) to reduce transient penalization of honest clients.

The extra server-side cost per round is one validation forward/backward pass; this cost scales linearly with the number of participating clients and is small compared to local training. For stable operation we recommend choosing the meta learning rate \(\eta\) smaller than the typical client learning rate and tuning it on the meta-validation set.
\vspace{-12ex}
\section{Experimental Evaluation}\label{experiments}
We evaluate FedAOT on all benchmark datasets using a federated learning simulation with 20 to 100 clients, depending on the experiment. To assess robustness under different adversarial levels, we introduce Byzantine clients ranging from 20\% to 90\% of the total population (A20 to A90). Each client trains a CNN and FNN hybrid model locally using the Adam optimizer with a learning rate of 0.001 and a batch size of 32. The Experiments were conducted in a Linux-based environment (Kaggle Notebooks) equipped with two NVIDIA T4 GPUs. All federated simulations were implemented using the Flower framework with PyTorch, ensuring consistent execution flow of client-server communication and reproducible execution across runs.

Since the attacks considered are untargeted, malicious clients in the label-flipping scenario modify the entire label set using the rule $(\text{label} + 1) \bmod 10$. This alteration affects the complete class distribution; therefore, we report overall classification accuracy instead of class-wise accuracy.

We compare FedAOT with three standard aggregation methods: FedAvg\cite{mcmahan2017}, FoolsGold\cite{fung2020limitations}, and GeoMed\cite{yin2018byzantine}. Performance is evaluated on the test set using both classification accuracy and the F1 score under each attack intensity.
\vspace{-10ex}
\subsection{Dataset} The proposed method is evaluated using benchmark datasets commonly used in FL research, such as MNIST\cite{lecun2002}, KMNIST\cite{clanuwat2018}, and FashionMNIST\cite{xiao2017}. These datasets provide a diverse set of tasks suitable for assessing the robustness of the aggregation method under various attack scenarios. 
\vspace{-10ex}
\subsection{Results}\label{sec6}
In this section, we discussed three primary aspects to evaluate the effectiveness and robustness of the proposed FedAOT aggregation method:\\
\\
\textit{\textbf{i. FedAOT Performance Against Untargeted Label-Flipping Attacks\\}}
To assess the standalone effectiveness of FedAOT under varying attack intensities, we evaluate its performance on MNIST, KMNIST, and FashionMNIST across attack levels ranging from A20 to A90. The corresponding results are presented in Tables~\ref{tab1}, \ref{tab2}, and \ref{tab3}.

\begingroup
\setlength{\textfloatsep}{0.5pt}
\begin{table}[H]
\centering
\caption{Performance Comparison Under Different Attack Intensities}
\label{tab:all}
\vspace{2mm}

\begin{subtable}{\textwidth}
\centering
\caption{MNIST}
\label{tab1}
\begin{tabular}{@{}llllllllll@{}}
\toprule
A20 & A30 & A40 & A50 & A60 & A70 & A80 & A85 & A90 \\
\midrule
98.78 & 98.43 & 97.71 & 97.85 & 97.31 & 98.19 & 97.08 & 98.11 & 97.77 \\
98.77 & 98.43 & 97.73 & 97.87 & 97.33 & 98.20 & 97.08 & 98.12 & 97.75 \\
\bottomrule
\end{tabular}
\end{subtable}

\vspace{3mm}
\begin{subtable}{\textwidth}
\centering
\caption{KMNIST}
\label{tab2}
\begin{tabular}{@{}llllllllll@{}}
\toprule
A20 & A30 & A40 & A50 & A60 & A70 & A80 & A85 & A90 \\
\midrule
94.84 & 93.85 & 92.46 & 90.60 & 91.38 & 88.22 & 92.24 & 88.70 & 91.66 \\
94.83 & 93.84 & 92.45 & 90.59 & 91.38 & 88.21 & 92.22 & 89.91 & 91.64 \\
\bottomrule
\end{tabular}
\end{subtable}

\vspace{3mm}

\begin{subtable}{\textwidth}
\centering
\caption{FMNIST}
\label{tab3}
\begin{tabular}{@{}llllllllll@{}}
\toprule
A20 & A30 & A40 & A50 & A60 & A70 & A80 & A85 & A90 \\
\midrule
89.53 & 89.31 & 89.27 & 88.83 & 88.78 & 88.90 & 87.97 & 88.74 & 88.46 \\
89.54 & 89.20 & 89.22 & 88.72 & 88.72 & 88.88 & 88.09 & 88.76 & 88.44 \\
\bottomrule
\end{tabular}
\end{subtable}
\end{table}
\endgroup
The results clearly demonstrate the robustness of FedAOT across all three datasets under untargeted label-flipping attacks. Accuracy remains consistently high even when up to 90\% of the participating clients are malicious, and both accuracy and F1 score show only minor variations across attack levels. This stability indicates that FedAOT successfully neutralizes the influence of poisoned updates.
\\
A notable observation is that the algorithm maintains performance regardless of the baseline difficulty of the dataset. For MNIST, the accuracy decreases only marginally from 98.78\% (A20) to 97.77\% (A90). KMNIST and FMNIST follow the same pattern, confirming that FedAOT effectively preserves model quality even in more challenging scenarios. The primary limiting factor becomes the model's inherent performance rather than the attack strength.
\\
\\
FedAOT’s resilience can be attributed to its adaptive weighting mechanism, which downweights adversarial updates without being influenced by their sheer volume. This property ensures stable global model updates even under extreme attack conditions, making FedAOT highly suitable for real-world federated learning deployments where large-scale Byzantine behavior may occur.\\
\\
\textit{\textbf{ii. Comparative performance of FedAOT against existing robust aggregation algorithms under Byzantine attacks\\}}
We evaluate FedAOT against FoolsGold, GeoMed, and FedAvg under increasing attack intensities (A20–A70).

\begin{figure}[ht]
    \centering
    \includegraphics[width=12cm]{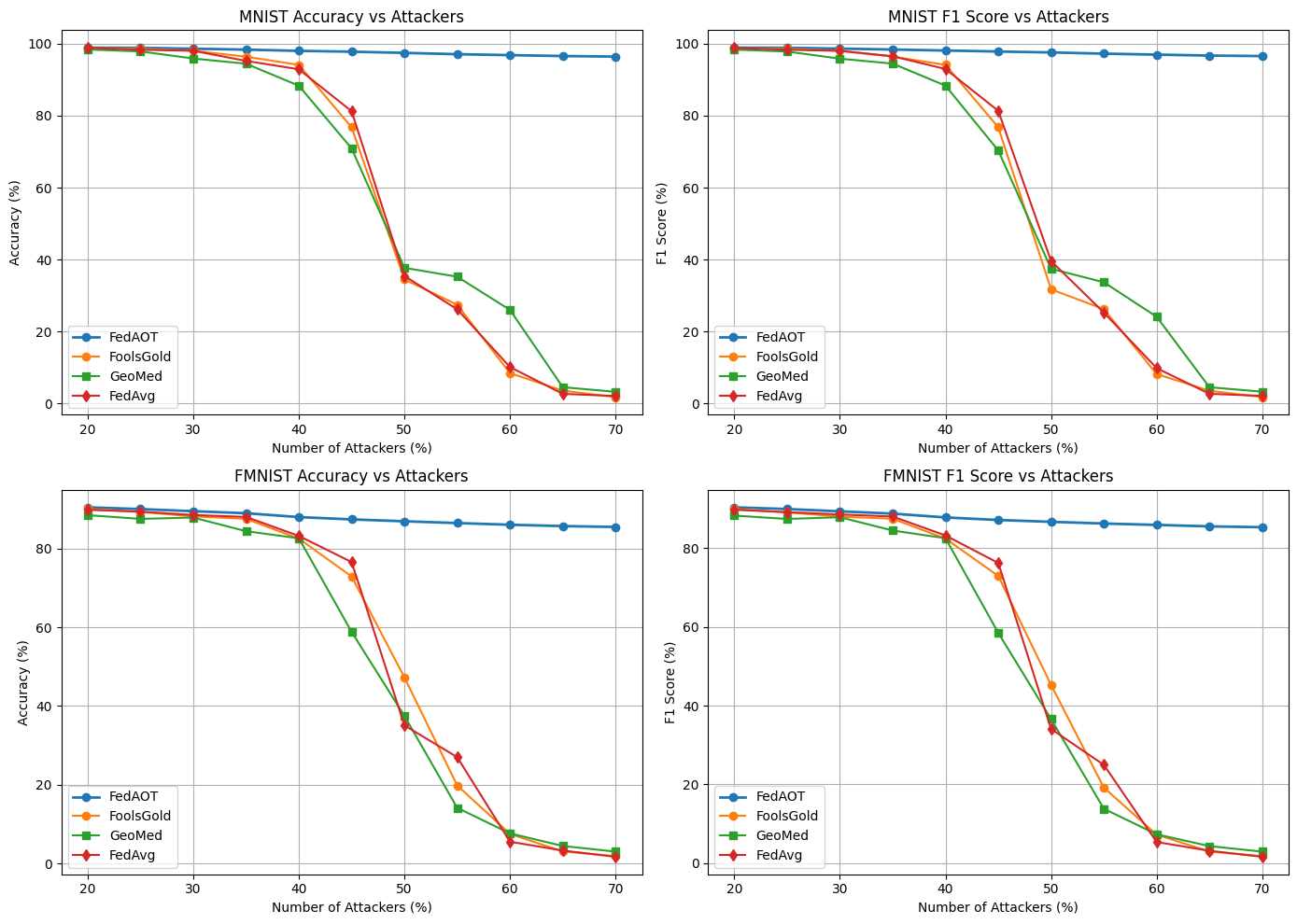}
    \caption{Comparison of aggregation methods under different attack intensities.}
    \label{fig:comparison}
\end{figure}

\noindent\textbf{Note.}
Krum and its variants were excluded due to extreme instability in untargeted poisoning scenarios. Since they select only one or a few client models per round, a single incorrect malicious selection collapses accuracy to near-zero, making them unreliable for this threat model.

\paragraph{\textbf{Performance at Low Attack Ratios (A20–A30):}}
At low attacker percentages, all baseline methods perform strongly, achieving over 98\% accuracy on MNIST and 88–90\% on FashionMNIST. Mild adversarial presence does not significantly disrupt learning.

\paragraph{\textbf{Performance Degradation with Increased Attacks (A35–A50)}:}
The baseline defenses begin to degrade noticeably as attack levels rise:
- At A40, FoolsGold and GeoMed decline to approximately 94\% and 88\% on MNIST, while FedAvg drops to 92.89\%.
- By A45–A50, performance collapses sharply: FoolsGold falls below 77\%, GeoMed reaches 70.9\%, and FedAvg drops to 81.24\%.
- On FashionMNIST, all methods fall below 76\% at A45, with severe instability at A50 (GeoMed: 37.49\%; FedAvg: 35.08\%).

\paragraph{\textbf{Complete Breakdown Beyond A50\\}}
- At A55, all baselines fail catastrophically, with F1 scores below 0.3 on MNIST and 0.2 on FashionMNIST.\\
- At A60–A65, FoolsGold and FedAvg operate near randomness, achieving only 3–4\% accuracy.\\
- At A70, all baseline defenses collapse fully, falling below 2\% accuracy.

In contrast, FedAOT preserves high accuracy across all attack intensities. As seen in Q1, the meta-layer effectively distinguishes honest from adversarial clients, sustaining stable performance even at A90.
\\
\\
\textit{\textbf{iii. Effectiveness of the FedAOT meta-layer in identifying and downweighting malicious clients}}\\
FedAOT utilizes adaptive weighting through learnable importance factors $k_i$ to distinguish between honest and malicious clients. This mechanism ensures that honest clients contribute meaningfully to the aggregation, while malicious clients are systematically downweighted. To demonstrate this, we analyze the distribution of $k$ - values across different attack intensities: A20, A50, A70, and A90.

\textbf{$k$ value Distribution Across Different Attack Levels:}\\
\includegraphics[width=12cm, height=12cm]{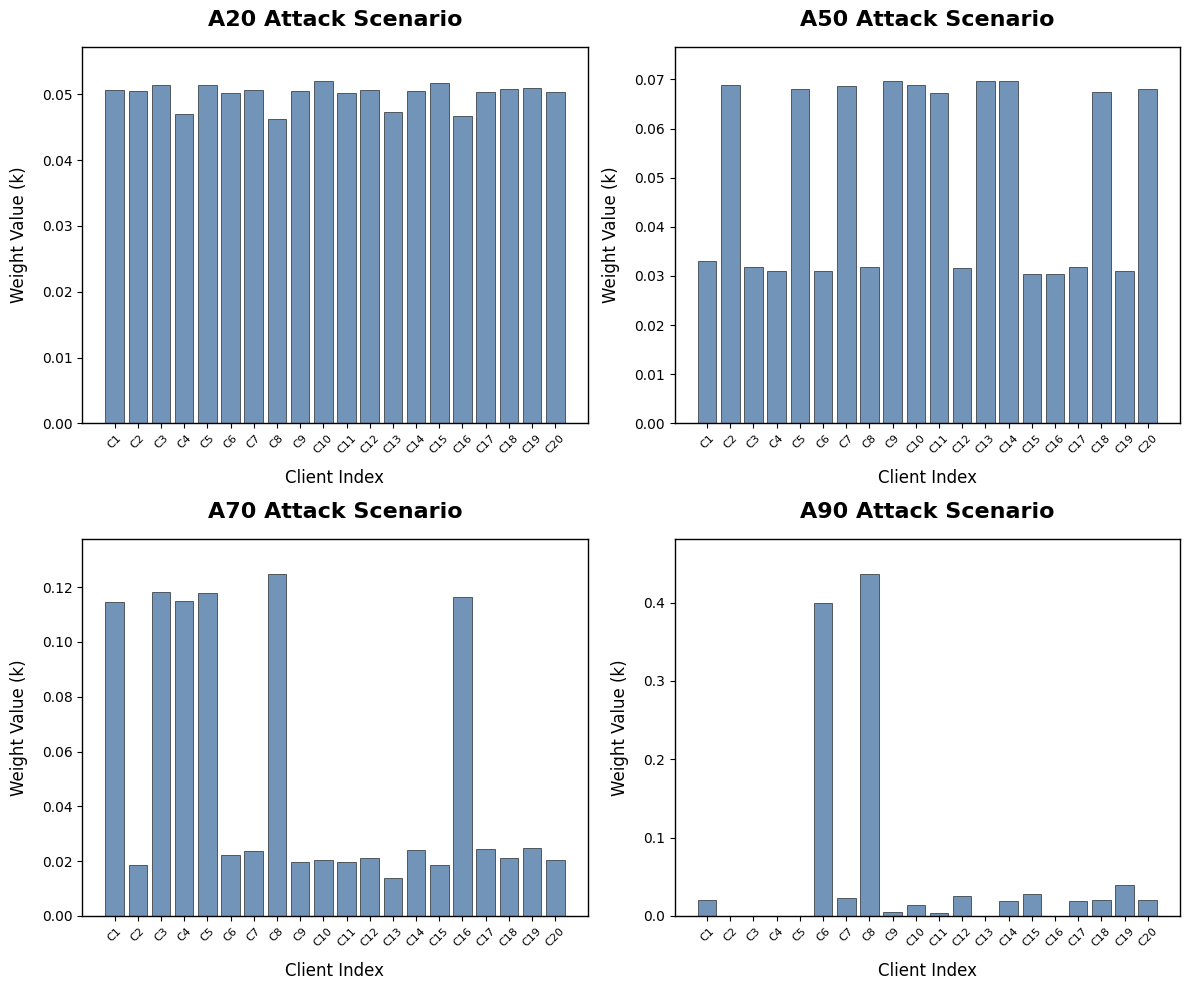}\\
For clarity and to avoid redundancy, we presented results using a 20-client system for FMNIST.

 Above, we present the distribution of $k$-values for 20 clients after 30 epochs under increasing 
attack intensities:

\begin{itemize}
	\item \textbf{A20:} The distribution is nearly uniform, indicating that all clients contribute equally, as the model has not detected significant adversarial influence.
	\item \textbf{A50:} The honest clients begin to receive higher weights, while malicious clients are downweighted, showing the algorithm's ability to differentiate between the two. 
	\item \textbf{A70:} The difference becomes more pronounced, with honest clients dominating the 
	weight distribution, ensuring that the model primarily learns from non-malicious 
	updates.
	\item \textbf{A90:} The system effectively isolates the two honest clients, giving them the 
	overwhelming majority of the weight, ensuring that aggregation is not compromised 
	by the large number of attackers.
\end{itemize}

 The gradual adaptation of $k$-values highlights two key strengths of FedAOT: 
 
 \begin{itemize}
 	\item \textbf{Robust Adaptive Filtering:} The algorithm effectively learns to separate honest and malicious clients over time, rather than making hard-coded assumptions about attacker behavior.  
 	\item \textbf{Stable Model Contribution:} Instead of relying on a single trusted client, FedAOT aggregates knowledge from all honest clients, ensuring long-term model stability. 
 	\item \textbf{Potential for Future Extensions:} Since the algorithm dynamically assigns lower weights to malicious clients, this technique could potentially be extended to defend against targeted poisoning attacks or more sophisticated Byzantine threats.  
 \end{itemize}
 This analysis reinforces that FedAOT is actively learning which clients should contribute more. 

\section{Conclusion and Future Scope}\label{conclusion}

The experimental results demonstrate that many existing robust aggregation strategies struggle when a large proportion of clients behave maliciously, largely because they fail to reliably distinguish honest updates from poisoned ones. Such approaches, particularly those relying on fixed rules or hard-selection, may degrade sharply under large-scale Byzantine activity.

FedAOT addresses these limitations through an adaptive importance weighting mechanism that adjusts each client's contribution via meta-layer optimization. This dynamic adjustment effectively suppresses adversarial updates while retaining the influence of honest clients, enabling the global model to remain stable even under extreme attack intensities. Across all datasets and attack levels, FedAOT consistently preserves high accuracy and robustness against untargeted label-flipping attacks.

Future extensions may include the stability and long-term behavior of the adaptive weighting method, particularly to understand the conditions under which the weights reliably converge during training. Also, developing more principled strategies for constructing meta-validation data—potentially through curated public datasets, synthetic data generation, or controlled volunteer contributions—depending on deployment requirements. Further evaluation under highly non-IID settings is also necessary to fully understand performance boundaries in heterogeneous environments.

\bibliographystyle{spphys} 
\bibliography{refer}
\end{document}